\documentclass{llncs}
\usepackage{makeidx}  

\usepackage{tikz}
\usepackage{pgfplots}

\usepackage{graphicx}
\usepackage{amsmath}
\usepackage{amssymb}
\usepackage{gensymb}

\newcommand{\figref}[1]{Figure \ref{#1}}
\newcommand{\tabref}[1]{Table \ref{#1}}

\usepackage[pagebackref=true,breaklinks=true,letterpaper=true,colorlinks,bookmarks=false]{hyperref}

\begin{document}

\mainmatter              
%

\title{Predicting the Next Best View for \\ 3D Mesh Refinement}
\titlerunning{Next Best View for 3D Mesh Refinement}  
%
\author{Luca Morreale \hspace{0.35cm}  Andrea Romanoni \hspace{0.35cm} Matteo Matteucci}
%
\authorrunning{Luca Morreale et al.} 
%
\tocauthor{Luca Morreale, Andrea Romanoni, Matteo Matteucci}
\institute{Politecnico di Milano, \\
\email{luca.morreale@mail.polimi.it}, \email{andrea.romanoni@polimi.it}, \email{matteo.matteucci@polimi.it}}

\maketitle              

\begin{abstract}
3D reconstruction is a core task in many applications such as robot navigation or sites inspections.
Finding the best poses to capture part of the scene is one of the most challenging topic that goes under the name of  Next Best View.
Recently, many volumetric methods have been proposed; they choose the Next Best View by reasoning over a 3D voxelized space and by finding which pose minimizes the uncertainty decoded into the voxels.
Such methods are effective, but they do not scale well since the underlaying representation requires a huge amount of memory.
In this paper we propose a novel mesh-based approach which focuses on the worst reconstructed region of the environment mesh.
We define a photo-consistent index to evaluate the 3D mesh accuracy, and an energy function over the worst regions of the mesh which takes into account the mutual parallax with respect to the previous cameras, the angle of incidence of the viewing ray to the surface and the visibility of the region.
We test our approach over a well known dataset and achieve state-of-the-art results.

\end{abstract}

\section{Introduction} \label{sec:introduction}
Two of the most challenging tasks for any robotic platform are the exploration and the mapping of unknown environments. 
When a surveying vehicle, such  as a drone, needs to autonomously recover the map of the environment, it has often time and power restrictions to fulfill.
A big issue is to incrementally look for the best pose to acquire a new measurement seeking a proper trade-off between the exploration of new areas and the improvement of the existing map reconstruction.
This problem is the so called Next Best View (NBV). 

NBV is usually addressed as an energy minimization/maximization problem. 
Most of the existing methods in 3D reconstruction rely on a volumetric reconstruction technique where each voxel is associated to its uncertainty and this information is used to compute the NBV.
From Multi-View Stereo literature it is well-founded that volumetric reconstructions are not able to scale well and they do not obtain accurate reconstructions with large datasets \cite{vu_et_al_2012}. 
Indeed on large scale problems the output of volumetric NBV algorithm has a coarse block-wise appearance.

Alternative reconstruction approaches base the estimation on 3D mesh representations; these methods do not need to store and reason about uncertainty on a volume, since they reason about a 2D manifold; therefore, they result to be more scalable. 
Moreover the underlaying representation can coincide with the output of accurate meshing algorithms such as \cite{vu_et_al_2012}. 
Mesh-based methods have been already used for exploration purposes: by navigating towards the boundary of the mesh a robot is able to explore unknown regions of the environment.
However, in mapping scenarios, the focus has to be more on  reconstruction accuracy, rather than exploration, therefore we cannot limit the algorithm to consider the boundary of the mesh, but a more comprehensive approach that takes into account the reconstruction accuracy, i.e., map refinement, is needed.

We propose a novel holistic approach to solve the NBV problem on 3D meshes which spots the worst regions of the mesh and looks for the best pose that improves those, specifically:
\begin{itemize}
\item a set of estimators to compute the accuracy of a given mesh, see \ref{sec:accuracy_estimation},
\item a novel mesh-based energy function to look for the NBV as a trade-off between exploration and refinement, see \ref{sec:energy}.

\end{itemize}

\section{Related Works} \label{sec:related_works}

The Next-Best View (NBV) has been addressed by many researcher and it is hard to define a fixed taxonomy. 
Since different domains have used different terms, in the following we adopt the classification proposed in the survey in \cite{scott2003view}. We refer the reader also to more recent surveys in \cite{chen2011active} and \cite{delmerico2017comparison}.

Model-based NBV algorithms assume a certain knowledge about the environment, for instance, Schmid \textit{et al.} \cite{schmid2012view} rely on a digital surface model (DSM) of the scene.
This assumption is restrictive and often not easy to fulfill, for this reason, in the following, we focus on non model-based algorithms that build a representation while they estimate the next best view.

One of the simplest method to estimate non-model-based  NBV has been proposed in \cite{nbv:wenhardt_active_reconstruction}. 
Among a fixed set of pre-computed poses, the authors choose  the pose that improves three statistics about the covariance matrices of the reconstructed 3D points: the determinant, the trace and the maximum eigenvalue. 
The main drawback of point-based methods, is that occlusions are not considered and the search space is limited to sampled poses.

A more robust and widespread approach relies on a volumetric representation of the scene by means of a 3D lattice of voxels, usually estimated by means of OctoMap \cite{hornung2013octomap}; each voxel collects the information needed to define the NBV. 
Connelly \cite{connolly1985determination} and Banta \cite{banta2000next} classify each voxel as occupied, freespace or unknown, and they base the next best view prediction on the number of unknown voxels that the camera would perceive.
Similarly, Yamauchi \cite{yamauchi1997frontier} counts the so called frontier voxels, which are the voxels between free and unknown space.
An extension of the frontier-based NBV was proposed by Bircher \textit{et al.} \cite{bircher2016receding}: they use a receding horizon NBV scheme to build a tree which is explored and exploited efficiently.

Previous methods focus on scene exploration, neglecting the problem of refining the model estimated on which we are more interested in a mapping scenario.
Vasquez \textit{et al.} \cite{nbv:vasquez_volumetric} plan the next view for a range sensor by relying on frontier voxels together with sensor overlapping, to also refine the model of the scene. 
This approach is suitable for time of flight or RGB-D cameras, but in case of RGB images the overlapping is not sufficient to evaluate if a new pose would improve the reconstructed model. 
Indeed, to ensure good parallax, we also need to consider the 3D position and orientation with respect to the other cameras.

While frontier-based methods are usually based on a counting metric, a different, probabilistic, approach to volumetric NBV has been proposed more recently in \cite{potthast2014probabilistic,kriegel2015efficient,isler2016information,nbv:mendez_next_stereo}. 
In \cite{potthast2014probabilistic} the authors collect the information about the unknown voxels, the visibility, and the occlusions in a Bayesian fashion.
Kriegel \textit{et al.} \cite{kriegel2015efficient} combine the use of a volumetric representation of the space to plan a collision free path and a mesh representation used to compute the region that requires exploration.
Isler \textit{et al.} \cite{isler2016information} propose a flexible framework that uses four information gain functions collected in the volumetric space.
Recently Mendez \textit{et al.} \cite{nbv:mendez_next_stereo} propose an efficient method to compute, among a set of candidate poses, both the NBV by considering only a part of the scene, and a method to add a further camera that forms a good stereo pairs with the NBV. 
The method has been extended in \cite{nbv:mendez_taking} to explore the continuous space of the scene, instead of limiting to a precomputed set of images.

Volumetric methods have shown to be effective and to properly deal with occlusions differently from point based ones. 
However they require the boundary of the space to be known in advance and, above all, their voxel-based representation does not scale with large scenes, as underlined in the Multi-View Stereo literature \cite{labatut2007efficient,vu_et_al_2012,li2016efficient,rec:efficient_multiview}.
Another class of NBV algorithms, named mesh-based and which directly relies on a 3D mesh reconstruction, has the advantage to directly outputs the model of the scene, in addition to scalability. 
Dunn and Frahm \cite{nbv:dunn_developing} define the 3D mesh reconstruction uncertainty and look for new poses which improve accuracy resolution and texture coherence.  
However, they require to estimate at each iteration a new 3D mesh model.
With a similar approach Mauro \textit{et al.} \cite{mauro2014unified} aggregate 2D saliency, 3D points uncertainty, and point density to define the NBV, however their method relies on a point cloud representation and cannot cope with occlusions.

Differently from the literature hereafter, in this paper, we propose a mesh-based approach to estimate the photometric uncertainty of a mesh and to find the next best view which improves the worst regions.
According to this metric the proposed approach is independent from the reconstruction algorithm adopted and it is able to predict the NBV such that it focuses and improves the worst part of the reconstruction.
Even if the method is particularly focused on mesh refinement, when the worst regions are located at the boundary of the mesh, it also explores new regions of the scene.


\section{Proposed Method} \label{sec:proposed_approach}
\begin{figure}[t]
	\centering
	\includegraphics[width=0.9\textwidth]{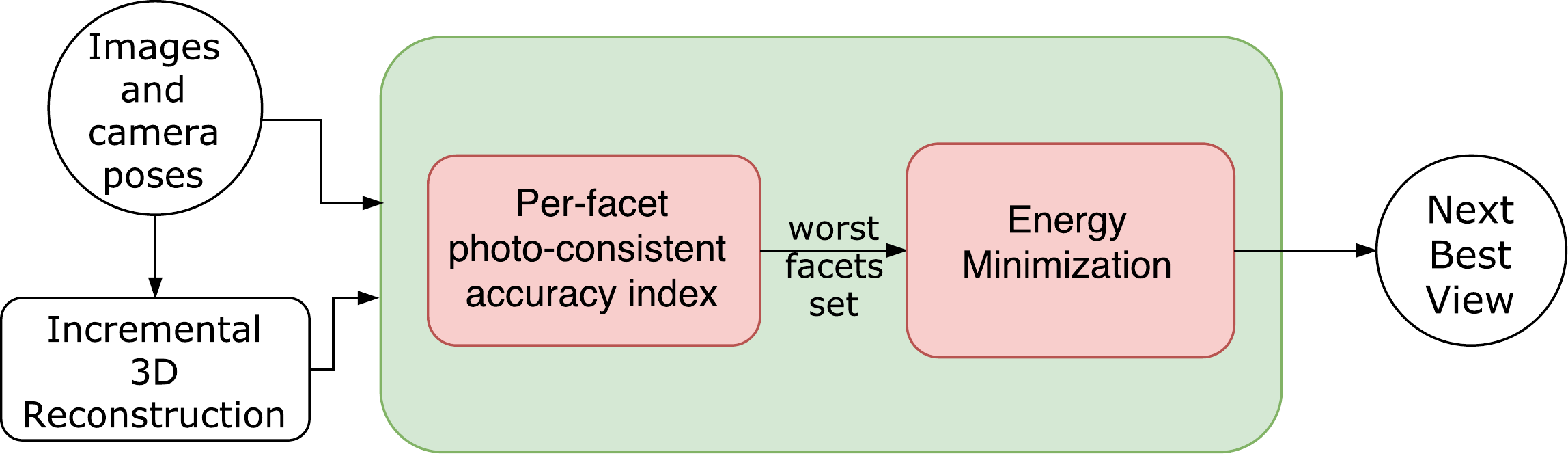}
	\caption{Architecture of the proposed Next Best View System}
	\label{fig:architecture}
\end{figure}

In this section we describe our  proposed NBV algorithm which relies on a mesh representation of the environment.
In the first step we estimate the 3D mesh uncertainty by computing a photo-consistent measure for each facet. 
Then, we select the worst facets and look for the pose which improves the accuracy of the reconstruction and, in case the selected facets lay near the boundary of the mesh, it also automatically improves the coverage.
In Figure \ref{fig:architecture} we illustrate the pipeline of the proposed method.

To keep  the mesh updated, we estimate new 3D points from each new pose, by means of the openMVG framework \cite{sfw:openMVG} and we apply the incremental  reconstruction algorithm proposed in  \cite{romanoni15b} and extended in \cite{rec:realtime_mesh}. 
The reconstruction algorithm builds a Delaunay Triangulation upon the 3D points and classifies the tetrahedra as free or occupied, the boundary between them represents the 3D mesh. As new points are estimated the triangulation and the reconstructed mesh are updated accordingly.

\subsection{Photo-consistent Reconstruction Index} \label{sec:accuracy_estimation}

As a first step we look for the regions of the model poorly reconstructed.
Since surfaces close to the actual surface project on the images in similar patches, and, in turn, surfaces far from it likely project in patches with different appearance, we adopt photo-consistency on those patches to check for poorly reconstructed regions.

Given a triangular mesh, we consider a facet $f$ and a pair of images $I_1$ and $I_2$ where the facet $f$ is visible; $f$ projects in two triangular patches $P_1\in I_1$ and $P_2 \in I_2$. 
We verify if $f$ is visible from an image $I$ by verifying that all its vertices are inside $I$ and that they are not occluded by other facets in the mesh.
To check the photo-consistency of $f$ with respect to $I_1$ and $I_2$, we compare  $P_1$ and $P_2$. 
Since their shape and dimension are arbitrary, we map them into a equilateral triangle with unitary sides; $P^f_1$ and $P^f_2$ become the mapped patches which we compare with a similarity measure $sim(P^f_1, P^f_2)$ and then we average among the whole set of images containing the same facet. This leads to the following formulation of the Photo-consistency Reconstruction Index (PRI) of facet $f$:
\begin{equation}
    PRI(f) = \frac{1}{|\mathcal{I}^f|} \sum_{P^f_i,  P^f_j \in \mathcal{I}^f}sim(P^f_i, P^f_j) ,
\end{equation}
computed between each couples, where  $\mathcal{I}^f$ is the pair of images where $f$ is visible.

We tested this estimator using as similarity measures the Sum of Squared Difference (SSD), i.e,
\begin{equation} \label{eq:ssd}
sim(P^f_1, P^f_2) = \sum_{x',y'} (P^f_1(x',y') - P^f_2(x',y'))^2 ,
\end{equation}
and  the Normalized Cross Correlation (NCC), i.e,
\begin{equation} \label{eq:ncc}
sim(P^f_1, P^f_2) = \frac{\sum_{x,y} (P^f_1(x,y)-\bar{P^f_1}) (I_2(x,y)-\bar{P^f_1})}       
{\sqrt{\sum_{x,y} (P^f_1(x,y)-\bar{P^f_1})^2  \sum_{x,y} (I_2(x,y)-\bar{P^f_1})^2}} ,
\end{equation}
where $\bar{P^f_1}$ and $\bar{P^f_2}$ represent the mean values of the patches.
We restricted the comparison of photo-consistency measures only to NCC and SSD since they are successfully adopted by the 3D reconstruction community, especially in multi-view stereo algorithms.
Nevertheless the approach can be used with any similarity metrics.

Concerning the scalability, our approach can clearly scale very well spatially since we use only a very small part of the image, thus even using a huge number of images the memory consumption is still low. On the other hand, the time complexity is $\mathcal{O}(n^2)$, where $n$ is the number of photos, thus it does not scale well. However, this issue can be solved considering a fixed number of views making the complexity constant, or, alternatively, computing the $PRI(f)$ only for the facets modified during the reconstruction.

\subsection{Next Best View} \label{sec:energy}
After we compute the per-facet accuracy, we select the $K=10$ facets with the lower $PRI$ and we collect them in the set $\mathcal{F}_w$. 
In the following we look for the Next Best View that is able to increase the accuracy of these facets.
To do this we propose a novel energy maximization formulation; it combines different contributions to cope with different requirements that a camera pose has to fulfill to improve both the existing reconstruction and to explore new parts of the environment.
In the following we refer to $\mathcal{V}_w$ as the set of vertices belonging to the facets in $\mathcal{F}_w$, and to $\mathcal{K}$ as penalization parameter. Specifically we experimentally choose  $\mathcal{K}=-10$ in order to enforce the presence of all terms of the energy function, since the violation of one of them would not result in a positive energy. 

\paragraph{Occlusion Term:}
The first term of the energy we want to minimize is named Occlusion Term, and it promotes the poses $P$ that sees the region we have to improve, i.e., the facets in $\mathcal{F}_w$ without occlusions.
We define this term: 
\begin{equation}\label{eq:occlusion}
O(P, v) =
\begin{cases}
1 & \text{if $v \in \mathcal{V}_w$ is not occluded,} \\
\mathcal{K} & \text{otherwise}  \\
\end{cases} .
\end{equation}
This term implicitly favors exploration. Indeed the facets in $\mathcal{F}_w$ are very likely located at the boundary of the mesh, and the Occlusion term leads the new view to focus on the region around $\mathcal{F}_w$. Furthermore, this formulation does not necessarily require to use GPU for the computation while computing the percentage of overlapping does.

\paragraph{Focus Term:}
The Focus Term represents the idea that the region around $\mathcal{F}_w$ preferably projects to the center of the image.
By favoring a projection around the center we also capture the surrounding regions and, since we have selected the worst reconstructed facets, we can fairly assume that also the nearby regions require an improvement.

To account for displacements with respect to the image center we weight the projection of a vertex $v \in \mathcal{V}_w$  with a 2D Gaussian distribution centered in the center of the image.
We formalize this term of the energy function as follows:
\begin{equation}\label{eq:image_projection}
\alpha =
-\frac{\left( {v^P_x - x_0 } \right)^2 }{2\sigma_x ^2} -\frac{\left( {v^P_y - y_0 } \right)^2}{2\sigma_y ^2},
\end{equation}
where $v^P_x$ and $v^P_y$ are the coordinates of $v$ projected on the camera $P$, and $(x_0, y_0)$ is the center of the image. 
We fix, $\sigma_x =  \frac{W}{3}$ and  $\sigma_y = \frac{H}{3}$ where $W$ and $H$ are the width and height of the image.
To take into account also the case in which $v$ projects outside the image, we rewrite Equation \eqref{eq:image_projection} as follows:
\begin{equation}\label{eq:projection_weight} 
F(P, v) =
\begin{cases}
\alpha & \text{if v is projected inside the image,} \\
\mathcal{K} & \text{otherwise} \\
\end{cases} .
\end{equation}

\paragraph{Parallax Term:}
With the Parallax Term we favor poses that capture the mesh with a significantly parallax with respect to the other images.
We base this term on the base-to-height (BH) constrain adopted in Aerial Photogrammetry \cite{egels2003digital} defined as $\frac{B}{H}>\delta$,
where $B$ is the baseline, i.e., the distance between two poses, and $H$ represents the distance between the  pose under evaluation and a point $v$, e.g., approximatively the distance between the robot and the surface, or the height of the drone in aerial surveys \cite{egels2003digital}.
Finally $\delta$ is a threshold that we experimentally fixed as $\delta=0.33$.
The Parallax term becomes:
\begin{equation}\label{eq:bd_constraint}
P(P, C) =
\begin{cases}
1 & \text{if $\frac{B}{H}>\delta$} \\
\mathcal{K} & \text{if $\frac{B}{H} \leq \delta$}  \\
\end{cases} .
\end{equation}
We compute this term with respect to each other camera $C$.

\paragraph{Incidence Term:}
The last term, named the Incident Term, of our energy function  encourages poses that observe the interested surface from an angle of incidence between $40\degree$ and $70\degree$.
The choice of such angles comes from the experience of the photogrammetric community as explained in \cite{photo:fraser1984network,photo:meixner2010characterizing,photo:nocerino2014accuracy}.
As presented in \cite{photo:meixner2010characterizing}, the smaller the incidence angle is the more distorted the information captured by an image.
Since we deal with angular quantities, we formalize it by means of a von Mises distribution from directional statistics:
\begin{equation}\label{eq:von_mises}
I(P, v) = log(\frac{e^{\kappa \cdot cos(x - \mu)}}{2 \pi I_0(\kappa)}),
\end{equation}
where $x$ represents the angle between the normal of the facet and the ray from facet barycenter to camera, $\mu$ is the angle with the highest probability (in our case $\mu=\frac{40+70}{2}$) and $\kappa$ is a measure of the concentration of the distribution, analogous to the inverse of the  variance. 
$I_0$ is the Bessel function of order zero:
\begin{equation}\label{eq:general_bessel}
I_0(k) = \sum_{m=0}^{\infty} \frac{(-1)^m}{m! \, \Gamma(m  + 1)} \left( \frac{\kappa}{2} \right)^{2m } .
\end{equation}

\paragraph{Next Best View Energy:}
We combine the previous four terms as it follows to obtain the energy function we want to minimize with the NBV algorithm:
\begin{equation}
\begin{split}
NBV(P, v) = &\; \mu_1 O(P, v) + \mu_2 F(P, v) + \\
&\; \mu_3 \sum_{c \in C} P(P, c) - \mu_4I(P, v) , 
\end{split}
\label{eq:energy_weighted}
\end{equation}
where $\mu_1$, $\mu_2$, $\mu_3$ and $\mu_4$ are the weights of the different terms, that must be experimentally tuned.
Then we define the energy over all vertices $v\in \mathcal{V}_w$ as:
\begin{equation}\label{eq:global_energy}
E(P) = \sum_{v\in \mathcal{V}_w} NBV(P, v).
\end{equation}

Although in the literature the energy is considered as cost, and thus the optimal pose must usually have the lowest energy, in the proposed approach terms used to compose the energy are higher for better poses. To frame the NBV as minimization we compute the negative of their sum.

\section{Experimental Validation} \label{sec:experiments}
To asses the quality of our method we tested it against both a custom synthetic dataset and the real dataset provided in \cite{ds:seitz_comparison}.
We run both experiments over a laptop equipped with Ubuntu 16.04 LST and Intel i7-7700HQ without the use of GPU. 

\begin{figure*}[t]
\centering
\begin{tabular}{cc} 
\includegraphics[width=0.4\columnwidth]{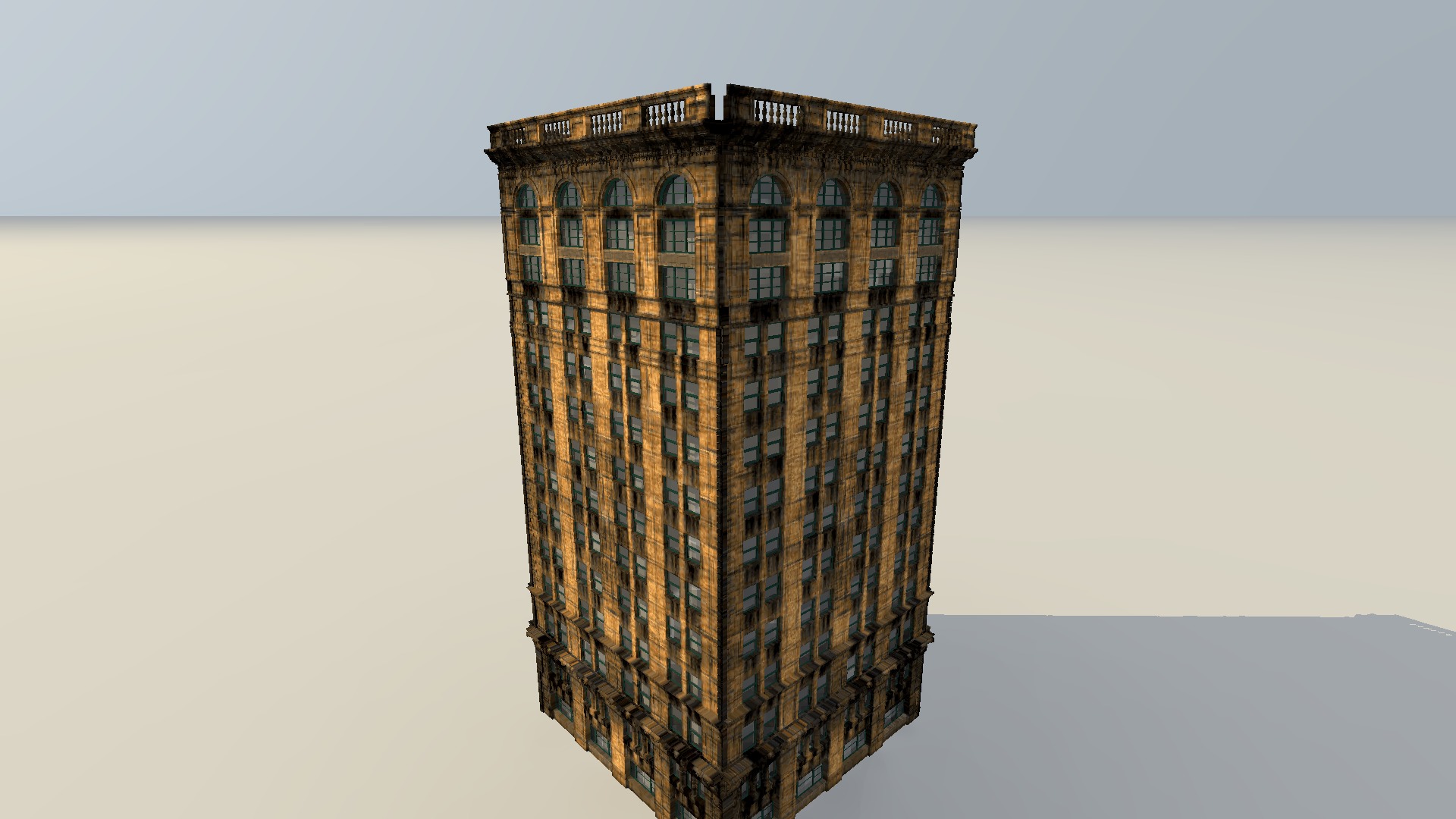} &
\includegraphics[width=0.4\columnwidth]{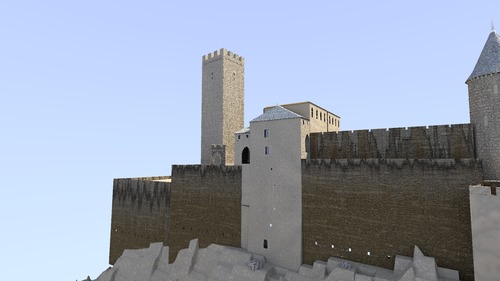} \\
\textbf{(a)} Building &
\textbf{(b)} Fortress \\
\includegraphics[width=0.4\columnwidth]{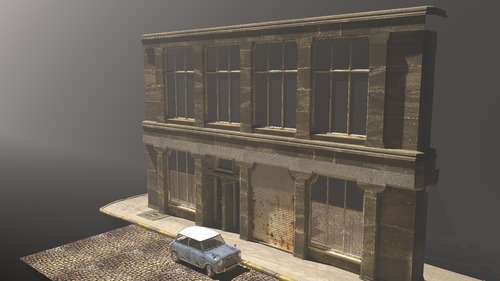} &
\includegraphics[width=0.4\columnwidth]{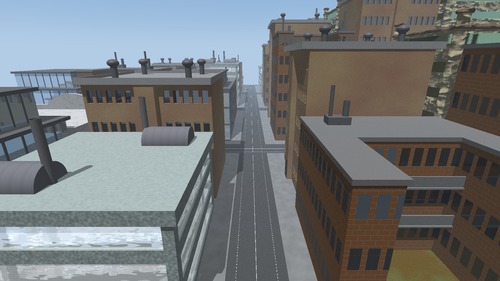} \\
\textbf{(c)} Car &
\textbf{(d)} City \\
\end{tabular}
\caption{Images from the four synthetic datasets.}
\label{fig:dataset}
\end{figure*}

Our synthetic dataset contains four photo-realistic environments generated by means of PovRay \cite{sfw:povray} and depicted in \figref{fig:dataset}; the big advantage of using such dataset is the possibility to capture an image from any point of view and so to simulate a real scenario where a robot, e.g. a drone, needs to navigate and explore the environment. 
Moreover the ground-truth model is available and therefore the reconstruction accuracy and completeness can be evaluated.

\begin{figure*}[t]
\centering
\begin{tabular}{ccc}
\includegraphics[width=0.3\columnwidth]{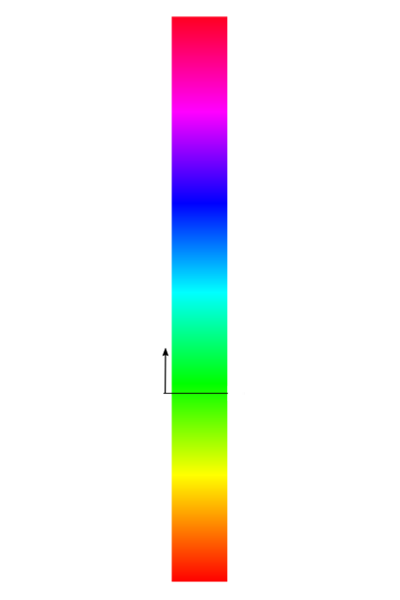} &
\includegraphics[width=0.3\columnwidth]{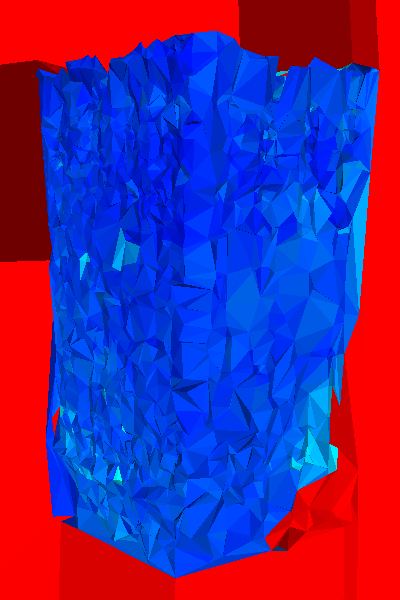} &
\includegraphics[width=0.3\columnwidth]{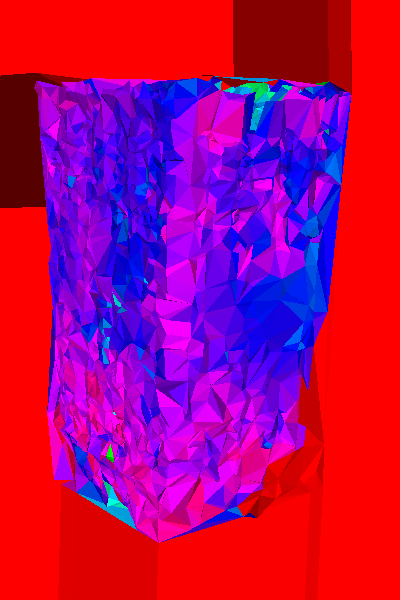} \\
\textbf{(a)} Color scale &
\textbf{(b)} NCC &
\textbf{(c)} SSD \\
\end{tabular}
\caption{Accuracy estimates in the synthetic dataset (first iteration). In (a) we show the color scale adopted, green corresponds to the optimal accuracy whilst red to the worst one.}
\label{fig:estimators}
\end{figure*}

The first step of our algorithm computes the accuracy for each facet; in \figref{fig:estimators} we illustrate the errors estimated on the reconstructed mesh with SSD and NCC as the color scale used, i.e., the hue of the HSV color space.
As shown in \figref{fig:estimators}(b), the NCC estimator provide a uniform underestimation of the accuracy. This estimation does not allow to distinguish clearly which areas need refinement. An appropriate estimation would show a lower accuracy around the corners and border, as in Figure \figref{fig:estimators}(c). Contrary, the SSD estimator does not provide a uniform estimation but properly indicate areas, like corners, as regions which need more refinement.

\begin{table}[t]
	\centering
 \setlength{\tabcolsep}{2px}
	\caption{Similarity measure computation time (on CPU).}
	\label{tab:time_estimators}
	\begin{tabular}{lc|cc|cc} 
        Similarity   & \# facets & \multicolumn{2}{c}{per-facet}  & \multicolumn{2}{c}{total} \\
        Measure  &  & NCC        & SSD               & NCC       & SSD \\
        \hline 
        Building            & 3440      & 0,043 ms & 0,043 ms & 150 ms & 150 ms \\
        Fortress            & 231       & 0,012 ms & 0,012 ms & 3 ms & 3 ms \\
        Car                & 314        & 0,038 ms & 0,035 ms & 12 ms & 11 ms \\
        City                & 43        & 0,651 ms & 0,302 ms & 28 ms & 13 ms \\
	\end{tabular}
\end{table}

\begin{figure}[t]
	\centering
	\begin{tabular}{cc}
		\includegraphics[width=0.4\columnwidth]{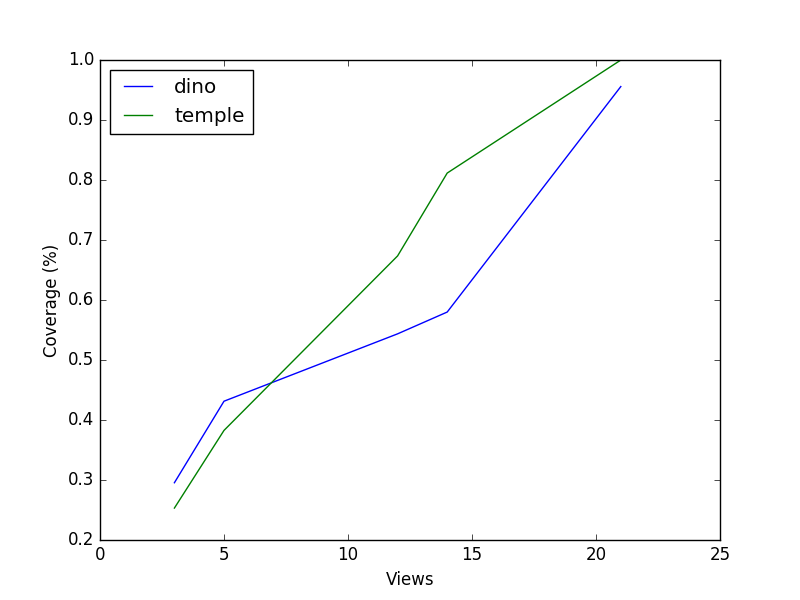} &
		\includegraphics[width=0.4\columnwidth]{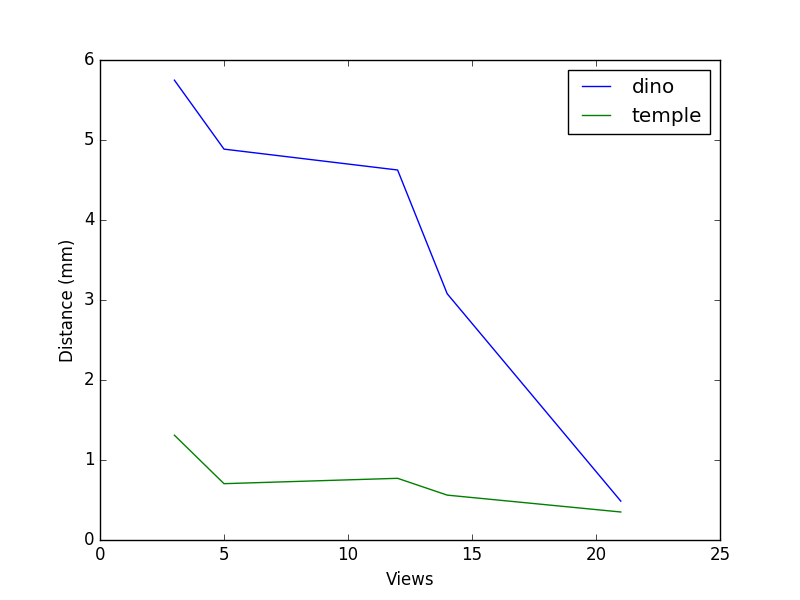} \\
		\textbf{(a)} Coverage &
		\textbf{(b)} Accuracy \\
	\end{tabular}
	 
	\caption{Reconstruction coverage (a) and accuracy(b) for the dino and temple datasets with different numbers of views.}
	\label{fig:coverage_accuracy}
\end{figure}

%

In \tabref{tab:time_estimators} we present the time required to estimate the NCC and SSD on the first reconstructed model, before estimating the first NBV. Even if we process the mesh on CPU, our method is able to rapidly estimate the accuracy. 


\begin{figure*}[tp]
\begin{tabular}{ccccc}
\includegraphics[width=0.175\textwidth]{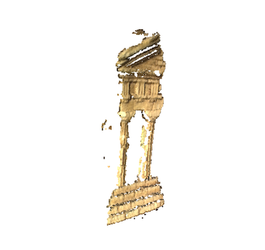} &
\includegraphics[width=0.175\textwidth]{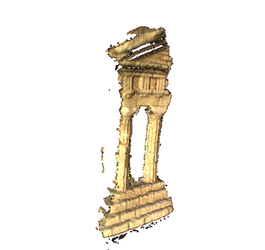} &
\includegraphics[width=0.175\textwidth]{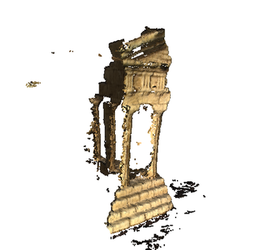} &
\includegraphics[width=0.175\textwidth]{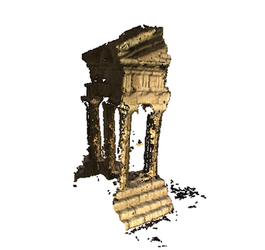} &
\includegraphics[width=0.175\textwidth]{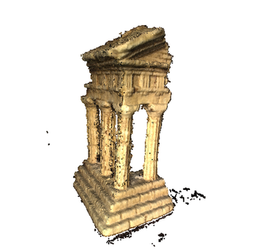} \\
\includegraphics[width=0.175\textwidth]{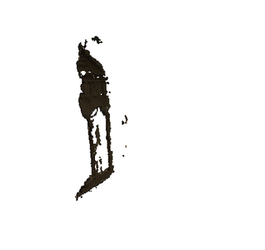} &
\includegraphics[width=0.175\textwidth]{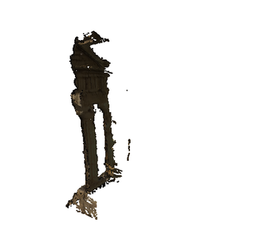} &
\includegraphics[width=0.175\textwidth]{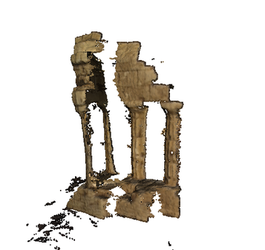} &
\includegraphics[width=0.175\textwidth]{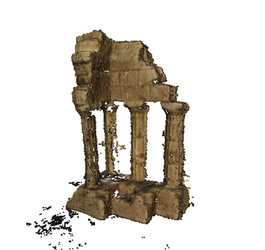} &
\includegraphics[width=0.175\textwidth]{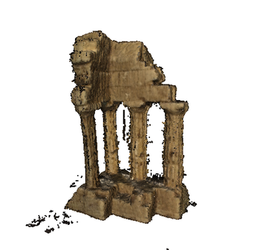} \\
{ 3 views} &
{ 5 views} &
{ 12 views} &
{ 14 views} &
{ 21 views} \\
\end{tabular}
\caption{Reconstructions of the temple with an incremental number of views.}
\label{fig:dino_increm_reconstruction}
\end{figure*}

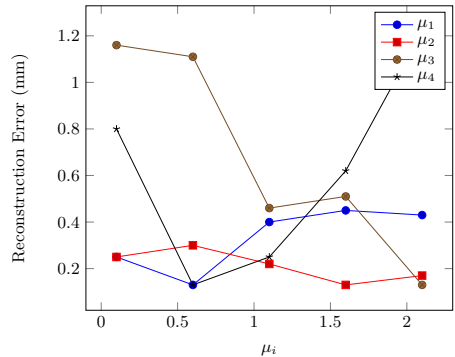
\begin{figure}[t]
\centering
\resizebox{0.5\textwidth}{!}{ %
\begin{tikzpicture}
    \begin{axis}[
        xlabel=\textsc{$\mu_i$},
        ylabel=Reconstruction Error (mm)
    ]
    \addplot plot coordinates {
    	(0.1, 0.25)
    	(0.6, 0.13)
        (1.1, 0.40)
        (1.6, 0.45)
        (2.1, 0.43)
    };
    \addplot plot coordinates {
    	(0.1, 0.25)
    	(0.6, 0.30)
        (1.1, 0.22)
        (1.6, 0.13)
        (2.1, 0.17)
    };
    \addplot plot coordinates {
    	(0.1, 1.16)
    	(0.6, 1.11)
        (1.1, 0.46)
        (1.6, 0.51)
        (2.1, 0.13)
    };
    \addplot plot coordinates {
    	(0.1, 0.8)
    	(0.6, 0.13)
        (1.1, 0.25)
        (1.6, 0.62)
        (2.1, 1.22)
    };
    
	\legend{$\mu_1$\\$\mu_2$\\$\mu_3$\\$\mu_4$\\}

    \end{axis}
\end{tikzpicture}
} %

\caption{The plot shows the reconstruction error at different values of the weights. Each curve represent a specific weight while the other are kept at their optimal value. Each weight has been tested in five different values: $0.1$ $0.6$ $1.1$ $1.6$ $2.1$.}
\label{fig:weights_plot}
\end{figure}

To prove the effectiveness of our approach we tested the whole system, i.e., accuracy estimation and energy function minimization, with a real dataset. Furthermore, given the considerations on the estimators, in the following we use SSD estimator. 
Concerning the dataset, we use the datasets dinoRing and templeRing presented in \cite{ds:seitz_comparison} (48 images for each sequence).
In this test we have a finite set of images and their relative position in the space.
Our goal is, starting from an initial set of images, incrementally select views and improve the reconstruction of the scene.

We bootstrap by computing the reconstruction of the scene using three views, then we use the NBV approach previously presented to incrementally select new images.
The system estimates which are the areas poorly reconstructed and uses the energy function to select a view, among those available, to enhance the mesh.
The view selected is the one having the highest energy score.

We run multiple time the selection process to tune the parameters $\mu_1$, $\mu_2$, $\mu_3$ and $\mu_4$ presented in Equation \eqref{eq:energy_weighted}.
The optimal weights are those providing the best mesh reconstruction using the lowest number of images.
Since no ground-truth is publicly available, we built a reference model with the reconstruction software Photoscan. 
For each mesh associated to a set of parameters we computed the accuracy of the 3D model generated with Photoscan and compared it with the reference model.
In Figure \ref{fig:coverage_accuracy} we report the reconstruction coverage and the accuracy obtained with the optimal parameters with respect to the real ground truth model; the proposed method is able to choose the convenient poses and rapidly improves both accuracy and coverage.
In Figure \ref{fig:dino_increm_reconstruction} we illustrate the meshes we reconstructed with the increasing number of views selected by our algorithm for the temple sequence.


The best results were obtained with $\mu_1 = 0.6$, $\mu_2=1.6$, $\mu_3 = 2.1$ and $\mu_4 = 0.6$.
To derive such values we systematically evaluated the reconstruction error, on dinoRing, changing the varying the weights and then test them with templeRing; we present the observed behavior in Figure \ref{fig:weights_plot}.
Our approach is able to effectively select a set of images, from a larger set, that properly represent the scene and allow an accurate reconstruction of it. 
\tabref{tab:results_compare} shows the accuracy and completeness for the dino dataset. Using less than half of the images in the dataset we are able to achieve comparable results with the other state of the art methods, reported in \cite{nbv:mendez_next_stereo}, that use only a subset of the images (as reported in \cite{nbv:mendez_next_stereo}), in particular with the volumetric method in \cite{nbv:mendez_next_stereo} although using $12\%$ less images.
We employed Photoscan\footnote{http://www.agisoft.com/} to generate the final model for both scenes.
In \tabref{tab:results_compare} we also reported the results on the temple dataset.
We compare only with \cite{nbv:jancosek_scalable}, since the other methods listed in \tabref{tab:results_compare} do not provide the results on the temple dataset. 
Our method is able to provide a significantly better coverage with comparable accuracy even using less images than state-of-the-art approaches.


\begin{table*}[t]
\footnotesize
	\caption{Evaluation of  different approaches.}
	\label{tab:results_compare}
	\centering
	\begin{tabular}{ccccccc|cc} 
   \hline 
   & & \multicolumn{5}{c}{dino} & \multicolumn{2}{c}{temple} \\
   & Thresholds& \cite{nbv:hornung_image} & \cite{nbv:hornung_image} & \cite{nbv:jancosek_scalable} & \cite{nbv:mendez_next_stereo} & Proposed& \cite{nbv:jancosek_scalable} & Proposed\\
   &  & Uniform & NBV & NBS & NBS & & & \\
   \hline \hline
   Num. Frames & - & 41 & 41 & unknown & 26 & 23 & unknown&23 \\
   \hline \hline
              & 80$\%$ & 0.64 & 0.59 & 0.64 & {0.53} & 0.61 & 0.51 & 0.57\\ 
   Error (mm) & 90$\%$ & 1.0 & 0.88 & 0.91 & {0.74} & 1.03 & 0.7 & 0.76 \\ 
              & 99$\%$ & 2.86 & 2.08 & 1.89 & {1.68} & 3.27 & 1.85 & 1.72 \\
   \hline \hline
                   & 0.75mm & 79.5 & 82.9 & 72.9 & {87.3} & 85.0 & 78.9 & 84.9 \\
   Coverage ($\%$) & 1.25mm & 90.2 & 93.0 & 73.8 & {96.4} & 93.6 & 78.9 & 95.0 \\
                   & 1.75mm & 94.3 & 96.9 & 73.9 & {98.4} & 97.2 & 78.9 & 97.5 \\
   \hline
	\end{tabular}
	
\end{table*}

\section{Conclusions and Future Works} \label{sec:conclusions}
In this paper we proposed a mesh-based algorithm to build a 3D mesh reconstruction by incrementally selecting the views that mostly improve the mesh.
Our approach estimates the reconstruction accuracy and then selects the views that enhance the worst regions maximizing a novel energy function.
We have demonstrated that our approach is able to achieve coverage and accuracy comparable to the state of the art, selecting incrementally images from a predefined set.
Our approach has the advantage to be independent from the reconstruction method used to build the model, as long as the output is a 3D mesh. 
As a future work, we plan to design a technique that, using the energy function proposed, defines the Next Best View in the  3D space instead of selecting a view from a set of precomputed poses.
Moreover we would test our approach with mesh-based methods as \cite{vu_et_al_2012}.

\bibliographystyle{splncs03}
\bibliography{egbib}

\begin{thebibliography}{10}
\providecommand{\url}[1]{\texttt{#1}}
\providecommand{\urlprefix}{URL }

\bibitem{banta2000next}
Banta, J.E., Wong, L., Dumont, C., Abidi, M.A.: A next-best-view system for
  autonomous 3-d object reconstruction. IEEE Transactions on Systems, Man, and
  Cybernetics-Part A: Systems and Humans  30(5),  589--598 (2000)

\bibitem{bircher2016receding}
Bircher, A., Kamel, M., Alexis, K., Oleynikova, H., Siegwart, R.: Receding
  horizon" next-best-view" planner for 3d exploration. In: Robotics and
  Automation (ICRA), 2016 IEEE International Conference on. pp. 1462--1468.
  IEEE (2016)

\bibitem{sfw:povray}
Buck, D.K., Collins, A.A.: {POV-Ray - The Persistence of Vision Raytracer}.
  \url{http://www.povray.org/}

\bibitem{chen2011active}
Chen, S., Li, Y., Kwok, N.M.: Active vision in robotic systems: A survey of
  recent developments. The International Journal of Robotics Research  30(11),
  1343--1377 (2011)

\bibitem{connolly1985determination}
Connolly, C.: The determination of next best views. In: Robotics and
  Automation. Proceedings. 1985 IEEE International Conference on. vol.~2, pp.
  432--435. IEEE (1985)

\bibitem{delmerico2017comparison}
Delmerico, J., Isler, S., Sabzevari, R., Scaramuzza, D.: A comparison of
  volumetric information gain metrics for active 3d object reconstruction.
  Autonomous Robots pp. 1--12 (2017)

\bibitem{nbv:dunn_developing}
Dunn, E., Van Den~Berg, J., Frahm, J.M.: Developing visual sensing strategies
  through next best view planning. In: Intelligent Robots and Systems, 2009.
  IROS 2009. IEEE/RSJ International Conference on. IEEE (2009)

\bibitem{egels2003digital}
Egels, Y., Kasser, M.: Digital photogrammetry. CRC Press (2003)

\bibitem{photo:fraser1984network}
Fraser, C.S.: Network design considerations for non-topographic photogrammetry.
  Photogrammetric Engineering and Remote Sensing  50(8),  1115--1126 (1984)

\bibitem{nbv:hornung_image}
Hornung, A., Zeng, B., Kobbelt, L.: Image selection for improved multi-view
  stereo. In: Computer Vision and Pattern Recognition, 2008. CVPR 2008. IEEE
  Conference on. pp. 1--8. IEEE (2008)

\bibitem{hornung2013octomap}
Hornung, A., Wurm, K.M., Bennewitz, M., Stachniss, C., Burgard, W.: Octomap: An
  efficient probabilistic 3d mapping framework based on octrees. Autonomous
  Robots  34(3),  189--206 (2013)

\bibitem{isler2016information}
Isler, S., Sabzevari, R., Delmerico, J., Scaramuzza, D.: An information gain
  formulation for active volumetric 3d reconstruction. In: Robotics and
  Automation (ICRA), 2016 IEEE International Conference on. pp. 3477--3484.
  IEEE (2016)

\bibitem{nbv:jancosek_scalable}
Jancosek, M., Shekhovtsov, A., Pajdla, T.: Scalable multi-view stereo. In:
  Computer Vision Workshops (ICCV Workshops), 2009 IEEE 12th International
  Conference on. pp. 1526--1533. IEEE (2009)

\bibitem{kriegel2015efficient}
Kriegel, S., Rink, C., Bodenm{\"u}ller, T., Suppa, M.: Efficient next-best-scan
  planning for autonomous 3d surface reconstruction of unknown objects. Journal
  of Real-Time Image Processing  10(4),  611--631 (2015)

\bibitem{labatut2007efficient}
Labatut, P., Pons, J.P., Keriven, R.: Efficient multi-view reconstruction of
  large-scale scenes using interest points, delaunay triangulation and graph
  cuts. In: Computer Vision, 2007. ICCV 2007. IEEE 11th International
  Conference on. pp. 1--8. IEEE (2007)

\bibitem{li2016efficient}
Li, S., Siu, S.Y., Fang, T., Quan, L.: Efficient multi-view surface refinement
  with adaptive resolution control. In: European Conference on Computer Vision.
  pp. 349--364. Springer (2016)

\bibitem{rec:efficient_multiview}
Li, S., Siu, S.Y., Fang, T., Quan, L.: Efficient multi-view surface refinement
  with adaptive resolution control. In: Leibe, B., Matas, J., Sebe, N.,
  Welling, M. (eds.) Computer Vision -- ECCV 2016. pp. 349--364. Springer
  International Publishing, Cham (2016)

\bibitem{mauro2014unified}
Mauro, M., Riemenschneider, H., Signoroni, A., Leonardi, R., Van~Gool, L.: A
  unified framework for content-aware view selection and planning through view
  importance. In: Proceedings BMVC 2014. pp. 1--11 (2014)

\bibitem{photo:meixner2010characterizing}
Meixner, P., Leberl, F.: Characterizing building facades from vertical aerial
  images. International Archives of Photogrammetry, Remote Sensing and Spatial
  Information Sciences  38(PART 3B),  98--103 (2010)

\bibitem{nbv:mendez_taking}
Mendez, O., Hadfield, S., Pugeault, N., Bowden, R.: Taking the scenic route to
  3d: Optimising reconstruction from moving cameras. In: Proceedings of the
  IEEE Conference on Computer Vision and Pattern Recognition (2017)

\bibitem{nbv:mendez_next_stereo}
Mendez, O., Hadfield, S., Pugeault, N., Bowden, R.: Next-best stereo: extending
  next best view optimisation for collaborative sensors. Proceedings of BMVC
  2016  (2016)

\bibitem{sfw:openMVG}
Moulon, P., Monasse, P., Marlet, R., Others: Openmvg. an open multiple view
  geometry library. \url{https://github.com/openMVG/openMVG}

\bibitem{photo:nocerino2014accuracy}
Nocerino, E., Menna, F., Remondino, F.: Accuracy of typical photogrammetric
  networks in cultural heritage 3d modeling projects. The International
  Archives of Photogrammetry, Remote Sensing and Spatial Information Sciences
  40(5),  465 (2014)

\bibitem{rec:realtime_mesh}
Piazza, E., Romanoni, A., Matteucci, M.: Real-time cpu-based large-scale 3d
  mesh reconstruction. In: Robotics and Automation (ICRA), 2018 IEEE
  International Conference on. IEEE (2018)

\bibitem{potthast2014probabilistic}
Potthast, C., Sukhatme, G.S.: A probabilistic framework for next best view
  estimation in a cluttered environment. Journal of Visual Communication and
  Image Representation  25(1),  148--164 (2014)

\bibitem{romanoni15b}
Romanoni, A., Matteucci, M.: Incremental reconstruction of urban environments
  by edge-points delaunay triangulation. In: IEEE/RSJ International Conference
  on Intelligent Robots and Systems (IROS). pp. 4473--4479. IEEE (2015)

\bibitem{schmid2012view}
Schmid, K., Hirschm{\"u}ller, H., D{\"o}mel, A., Grixa, I., Suppa, M.,
  Hirzinger, G.: View planning for multi-view stereo 3d reconstruction using an
  autonomous multicopter. Journal of Intelligent \& Robotic Systems  65(1),
  309--323 (2012)

\bibitem{scott2003view}
Scott, W.R., Roth, G., Rivest, J.F.: View planning for automated
  three-dimensional object reconstruction and inspection. ACM Computing Surveys
  (CSUR)  35(1),  64--96 (2003)

\bibitem{ds:seitz_comparison}
Seitz, S.M., Curless, B., Diebel, J., Scharstein, D., Szeliski, R.: A
  comparison and evaluation of multi-view stereo reconstruction algorithms. In:
  Computer vision and pattern recognition, 2006 IEEE Computer Society
  Conference on. vol.~1, pp. 519--528. IEEE (2006)

\bibitem{nbv:vasquez_volumetric}
Vasquez-Gomez, J.I., Sucar, L.E., Murrieta-Cid, R., Lopez-Damian, E.:
  Volumetric next-best-view planning for 3d object reconstruction with
  positioning error. International Journal of Advanced Robotic Systems  (2014)

\bibitem{vu_et_al_2012}
Vu, H.H., Labatut, P., Pons, J.P., Keriven, R.: High accuracy and
  visibility-consistent dense multiview stereo. Pattern Analysis and Machine
  Intelligence, IEEE Transactions on  34(5),  889--901 (2012)

\bibitem{nbv:wenhardt_active_reconstruction}
Wenhardt, S., Deutsch, B., Angelopoulou, E., Niemann, H.: Active visual object
  reconstruction using d-, e-, and t-optimal next best views. In: Computer
  Vision and Pattern Recognition, 2007. CVPR'07. IEEE Conference on. IEEE
  (2007)

\bibitem{yamauchi1997frontier}
Yamauchi, B.: A frontier-based approach for autonomous exploration. In:
  Computational Intelligence in Robotics and Automation, 1997. CIRA'97.,
  Proceedings., 1997 IEEE International Symposium on. pp. 146--151. IEEE (1997)

\end{thebibliography}

\end{document}